\documentclass[opre,sglanonrev]{informs4}
\usepackage{eqndefns-left} % For checking the display equation width and equation environment definitions %
\RequirePackage{tgtermes}
\RequirePackage{newtxtext}
\RequirePackage{newtxmath}
\RequirePackage{bm}
\RequirePackage{endnotes}
\usepackage{amsmath,amssymb,amsfonts}
\usepackage{textcomp}
\usepackage{algorithm}
\usepackage{algpseudocode}

\usepackage{cite}
\usepackage{subcaption}
\usepackage[table]{xcolor}
\usepackage{multirow}
\PassOptionsToPackage{pdftex}{graphicx}
\usepackage{booktabs}

\usepackage{listings}
\usepackage{rotating}

\OneAndAHalfSpacedXII % Current default line spacing
\usepackage{tikz}
\usepackage{natbib}
 \bibpunct[, ]{(}{)}{,}{a}{}{,}%
 \def\bibfont{\small}%
\EquationsNumberedThrough    % Default: (1), (2), ...
\TheoremsNumberedThrough     % Preferred (Theorem 1, Lemma 1, Theorem 2)
\ECRepeatTheorems  %  

\MANUSCRIPTNO{}

\begin{document}
%%%%%%%%%%%%%%%%

%\setcounter{page}{1}
%\DOI{10.1287/xxxx.0000.0000}%

\RUNAUTHOR{Karaahmetoglu and Kim}

% \RUNTITLE{Predictive Maintenance in Manufacturing}
\RUNTITLE{Collaborative Partition Optimization}

% \TITLE{Optimal Resource Allocation in Humanitarian Logistics: A Stochastic Programming Approach}
\TITLE{Robust Feasible Route Construction through Collaborative Partition Optimization}

\ARTICLEAUTHORS{%
%\AUTHOR{John Doe,\textsuperscript{a} Jane Smith,\textsuperscript{b}}
%\AFF{\textsuperscript{a}Department of Industrial Engineering, University of XYZ, \EMAIL{john.doe@xyz.edu; \textsuperscript{b}Department of Computer Science, University of ABC, \EMAIL{jane.smith@abc.edu}} 
\AUTHOR{Oguzhan Karaahmetoglu}
\AFF{Department of Electrical and Computer Engineering,
Carnegie Mellon University, \EMAIL{okaraahm@cmu.edu}}

\AUTHOR{Hyong Kim}
\AFF{Department of Electrical and Computer Engineering,
Carnegie Mellon University, \EMAIL{kim@ece.cmu.edu}}
} % end of the block

\ABSTRACT{%
    Large-scale Capacitated Vehicle Routing Problems (CVRPs) are commonly solved by partitioning customers into smaller routing problems that can be optimized independently. While this substantially reduces computational complexity, independently constructed routing solutions may leave some customer demand unserved even when sufficient resources exist elsewhere in the fleet. We present Collaborative Routing Constructors (CoRC), a routing framework that enables independently solved subproblems to exchange customers and vehicles during optimization rather than relying solely on a fixed partition or a subsequent global re-optimization stage. Computational experiments on AGS benchmark instances and synthetic instances containing up to 200,000 customers compare CoRC against independent routing, post-routing global re-optimization, and state-of-the-art, end-to-end routing frameworks. Across all evaluated partitioning strategies, CoRC consistently constructs feasible routing solutions where competing partition-based methods do not. Furthermore, it remains effective on problem instances for which the evaluated end-to-end routing frameworks did not produce solutions under the same computational budget. These results demonstrate that collaboration between routing subproblems provides a robust and scalable approach for feasible large-scale route construction.
}%

%Supplemental Material:
%Data Ethics & Reproducibility Note:

%\KEYWORDS{Stochastic programming, Decision support,Uncertainty, Disaster response, Optimization}

%\HISTORY{Received: Month DD, YYYY; Accepted: Month DD, YYYY; Published Online: Month DD, YYYY}

\maketitle
%%%%%%%%%%%%%%%%%%%%%%%%%%%%%%%%%%%%%%%%%%%%%%%%%%%%%%%%%%%%%%%%%%%%%%

% Text of your paper here
% \vspace{0.5em}
% \noindent Subject Classification
% \begin{itemize}
%     \item Transportation: Vehicle routing - Feasible large-scale CVRP construction
%     \item Networks/graphs: Applications - Collaborative decomposition for vehicle routing
%     \item Programming: Heuristic - Collaborative partition optimization
% \end{itemize}

% \noindent Area of Review:
% Optimization

\section{Introduction}\label{sec:Intro}
    We consider the Capacitated Vehicle Routing Problem (CVRP) with both homogeneous \citep{dantzig1959truck} and heterogeneous vehicle fleets \citep{golden1984fleet,toth2014vrp}. Given a set of customers with known demands and a fleet of capacitated vehicles, the objective is to build sequence of customer visits that serve every customer while minimizing the total routing cost without violating vehicle-capacity constraints. The CVRP is one of the fundamental optimization problems in transportation and logistics, with applications including freight distribution \citep{crainic1997planning}, parcel and last-mile delivery \citep{boysen2020last}, humanitarian logistics \citep{campbell2008routing}, and urban service operations such as waste collection and street maintenance \citep{yamin2024reliable,anily1993two}.

    The CVRP is NP-hard \citep{lenstra1981complexity}. Consequently, the literature has developed a broad range of exact algorithms, heuristics, metaheuristics, and decomposition techniques for obtaining high-quality routing solutions \citep{gillett1974heuristic, clarke1964scheduling, pisinger2018large}. Among these, decomposition-based approaches have become an important component of modern vehicle-routing heuristics, particularly for large routing instances \citep{santini2023decomposition}.

    One prominent decomposition paradigm is the cluster-first route-second (CFRS) framework \citep{gillett1974heuristic, jingjing2022cluster}. In this framework, the original routing instance is partitioned into multiple routing subproblems, where each subproblem consists of a subset of customers together with an assigned subset of vehicles. Each subproblem therefore defines a smaller CVRP that can subsequently be routed independently. Thus, a routing algorithm is applied to each subproblem to construct local vehicle routes \citep{kerscher2024spatial, pisinger2018large}. Finally, the routes obtained from all subproblems are then merged to produce a solution for the original large-scale instance. This separation between partitioning and routing naturally raises the question of how routing should be performed after the decomposition has been obtained. To this end, we discuss the proposed methods for this paradigm in the next subsection.

\subsection{Prior Art}
    Early CFRS methods focused primarily on geometric partitioning. Representative examples include the sweep algorithm \citep{gillett1974heuristic} and centroid-based clustering approaches \citep{abdellaoui2024towards}, which partition customers according to their spatial locations before building routes independently within each partition. Numerous subsequent methods have continued to build upon geometric partitioning principles while proposing alternative clustering procedures \citep{cakir2015revisiting, gaon2025optimizing}.

    Subsequent work incorporated additional operational information into the partitioning stage. Rather than relying exclusively on customer locations, these methods also consider quantities such as customer demand, vehicle capacities, or routing-related workload measures when constructing partitions \citep{alesiani2022constrained, linfati2022mathematical, jingjing2022cluster}. More recently, learning-based approaches have used optimization feedback to guide decomposition decisions through reinforcement learning, neural combinatorial optimization, and large language models \citep{li2022overview, chin2026neural, da2026large, xiu2026large, thind2025optimai}. These modern approaches extend the former by exploiting the patterns collected in the historical data by estimating a policy via empirical risk minimization \citep{vapnik1991principles}.

    After the decomposition has been constructed, the most common routing strategy is to solve each routing subproblem independently before combining the resulting routes into a complete solution \citep{gillett1974heuristic, battarra2014exact}. Under this strategy, each cluster is processed via a routing solver using only the customers and vehicles assigned to its own subproblem.

    Another line of studies perform an additional optimization stage after the independently produced routes have been merged \citep{kerscher2024spatial}. These approaches apply global improvement procedures, such as adaptive large neighborhood search \citep{ropke2006adaptive} or decompose-route-improve strategies, to refine the combined routing solution. By optimizing the merged routing plan together with the combined customer and vehicle assignments, these methods provide an additional opportunity to improve the solution obtained from independent route construction. This aims to reduce routing cost or incorporate customers that remained unserved after the independent routing stage.

    Existing routing paradigms either preserve the initial decomposition throughout route construction or introduce coordination only through a subsequent global re-optimization stage. However, route construction itself generates optimization information that is unavailable during the initial decomposition, including unserved demand, unused vehicles, and remaining vehicle capacity. These quantities characterize the current routing state of each subproblem and may indicate opportunities to revise customer and vehicle assignments before the routing process is complete.

    Motivated by this observation, we investigate an alternative organization of the routing stage. Rather than restricting coordination to the initial decomposition or postponing it until a subsequent global optimization stage, we introduce an intermediate collaboration stage during route construction. In this scheme, routing subproblems exchange summaries of their local routing outcomes, enabling customer transfers, vehicle transfers, and partition merges before the final solution is assembled. Because the collaboration mechanism operates on an existing initial decomposition, it is complementary to geometric, operational, and learning-based partitioning methods.

\section{Problem Definition} \label{sec:problem_formulation}
    Our method builds upon the classical CVRP and its decomposition under a cluster-first route-second framework. Accordingly, we first formulate the underlying routing problem, then define the corresponding decomposition, and finally introduce a dynamic extension that allows routing subproblems to interact during routing.

    \begin{figure*}[h]
        \centering
        \begin{subfigure}[t]{0.32\textwidth}
            \centering
            \includegraphics[width=\linewidth]{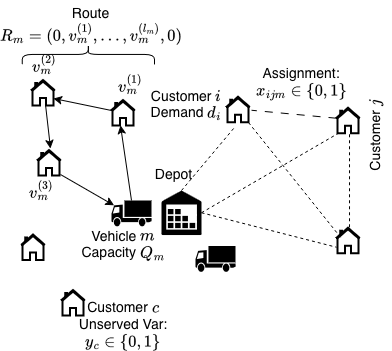}
            \caption{Capacitated Vehicle Routing Problem formulation.}
            \label{fig:cvrp_formulation}
        \end{subfigure}
        \hfill
        \begin{subfigure}[t]{0.32\textwidth}
            \centering
            \includegraphics[width=\linewidth]{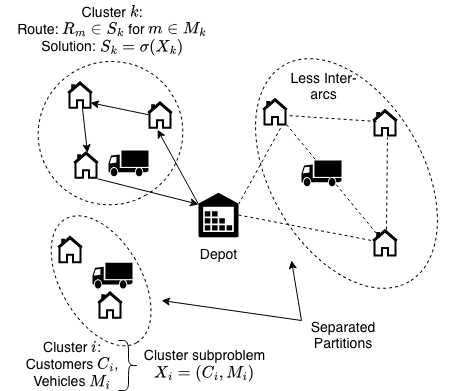}
            \caption{Static Cluster-First Route-Second (CFRS) formulation.}
            \label{fig:cfrs_static}
        \end{subfigure}
        \hfill
        \begin{subfigure}[t]{0.32\textwidth}
            \centering
            \includegraphics[width=\linewidth]{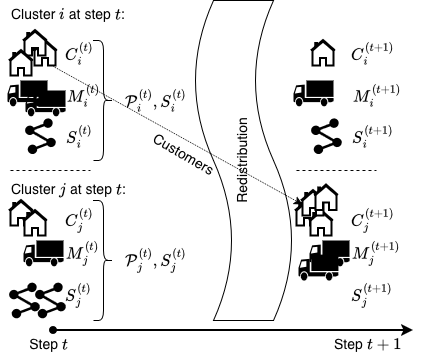}
            \caption{Evolving CFRS formulation with incumbent solutions.}
            \label{fig:cfrs_dynamic}
        \end{subfigure}
    
        \caption{
        Mathematical progression of the presented formulation. (a) CVRP formulation defining routes and routing solutions. (b) Conventional CFRS formulation, where a fixed decomposition $\mathcal{P}$ induces independent routing subproblems whose incumbent solutions $\mathcal{S}$ are combined into a global solution. (c) Evolving CFRS formulation, where the decomposition and incumbent solutions evolve over optimization steps through redistribution operations while preserving unaffected incumbent routes.
        }
        \label{fig:formulation_overview}
    \end{figure*}

    \subsection{Capacitated Vehicle Routing with Unsplittable Demands}
        In this section, we introduce the formulation and the notation used throughout the paper along with the CVRP objective and constraints. A simple case is illustrated in Figure~\ref{fig:cvrp_formulation}. Let $C$ denote the set of customers, $M$ the set of vehicles, and $L=C\cup\{0\}$ the set of all locations including the depot. Each customer $c\in C$ has demand $d_c$, each vehicle $m\in M$ has capacity $Q_m$, and the travel cost between locations $i,j\in L$ is given by $t_{ij}$.
        
        \begin{definition}[Route]
            A route for vehicle $m\in M$ is an ordered sequence of locations $ R_m=(0,v_m^{(1)},\ldots,v_m^{(l_m)},0), $ where $ l_m $ is the route length of vehicle $ m $ and $v_i\in C$ are distinct customers visited by vehicle $m$, the route begins and ends at the depot, and the total demand served along the route does not exceed the vehicle capacity, $ \sum_{c\in R_m} d_c \le Q_m$.
        \end{definition}
        
        \begin{definition}[Routing Solution]
            A routing solution is a collection of routes $ S=\{R_m\}_{m\in M} $, containing at most one route for each vehicle in the fleet.
        \end{definition}
        
        Throughout this paper, we distinguish several types of routing solutions.
        
        \begin{definition}[Solution Terminology]
            Let $S$ denote a routing solution.
            \begin{itemize}
                \item A \emph{feasible solution} satisfies all routing constraints of the optimization model but may leave some customer demand unserved.
                \item A \emph{full solution} is a feasible solution in which every customer is served.
            \end{itemize}
        \end{definition}
        
        To accommodate routing instances in which the available vehicle resources are insufficient to serve all customer demand, or intermediate optimization stages in which a complete solution has not yet been constructed, we permit customers to remain temporarily unserved. Let $x_{ijm}$ be a binary variable indicating whether vehicle $m$ traverses arc $(i,j)$, let $z_{cm}$ be a binary variable indicating whether customer $c$ is assigned to vehicle $m$, and let $y_c$ indicate whether customer $c$ remains unserved. We consider the following objective:
        \begin{equation}
            \min \sum_{m\in M}\sum_{i\in L}\sum_{j\in L} t_{ij}x_{ijm} + \rho\sum_{c\in C}d_cy_c,
        \end{equation}
        where $ \rho \in \mathbb{R}^+ $ is a penalty coefficient associated with unserved demand. The objective minimizes total travel cost while penalizing customer demand that is not served.
        
        The routing solution is subject to standard assignment, flow-conservation, and vehicle-capacity constraints \citep{laporte2009fifty}. Each customer must either be visited exactly once or remain unserved,
        \begin{equation}
            \sum_{m\in M}\sum_{j\in L} x_{cjm} + y_c = 1, \qquad \forall c\in C.
        \end{equation}
        
        Flow conservation requires that every vehicle entering a customer location must also depart from that location,
        \begin{equation}
            \sum_{j\in L} x_{ijm} = \sum_{j\in L} x_{jim}, \qquad \forall i\in C,\; m\in M.
        \end{equation}
        
        Vehicle capacities are enforced through
        \begin{equation}
            \sum_{c\in C} d_cz_{cm} \le Q_m, \qquad \forall m\in M.
        \end{equation}
        
        Additional routing constraints, including depot connectivity, subtour elimination, and binary-domain restrictions, follow the standard CVRP formulation and are omitted for brevity.

    \subsection{Cluster-First Route-Second Formulation}
        We next formalize the cluster-first route-second (CFRS) framework used throughout this paper. We begin with the conventional static formulation, in which the decomposition remains fixed during route construction. We then extend this formulation by allowing customer and vehicle assignments to evolve over successive optimization rounds, providing the mathematical basis for the collaborative routing method developed in the following sections.
        
        \subsubsection{Static Formulation}
            A cluster-first route-second (CFRS) decomposition partitions a CVRP instance into a collection of smaller routing subproblems as shown in Figure~\ref{fig:cfrs_static}. Let
            \begin{equation}
                \mathcal{P}=\left\{(C_1,M_1),(C_2,M_2),\dots,(C_K,M_K)\right\}
            \end{equation}
            denote a decomposition of the customer set $C$ and vehicle fleet $M$, where each routing subproblem consists of a customer subset $C_k\subseteq C$ and an associated vehicle subset $M_k\subseteq M$. The customer and vehicle assignments form disjoint partitions,
            \begin{align}
                C_i\cap C_j &= \emptyset,
                & \forall i\neq j,\\
                \bigcup_{k=1}^{K}C_k &= C,\\[0.3em]
                M_i\cap M_j &= \emptyset, & \forall i\neq j,\\
                \bigcup_{k=1}^{K}M_k &= M.
            \end{align}
            
            Each partition defines an independent routing subproblem $ X_k=(C_k,M_k) $, which is solved using a routing procedure
            \begin{equation}
                S_k=\sigma(X_k)=\{R_m\}_{m\in M_k},
            \end{equation}
            where $S_k$ denotes the incumbent solution associated with subproblem $X_k$. Note that the routes contained in the set are only for the vehicles of the partition $ k $. The collection of incumbent solutions is given by $ \mathcal{S}=\left\{S_1,\ldots,S_K\right\} $.
            
            \begin{definition}[Subproblem Solution]
                The incumbent solution $S_k\in\mathcal{S}$ associated with routing subproblem $X_k$ is referred to as a \emph{subproblem solution}.
            \end{definition}
            
            \begin{definition}[Global Solution]
                The \emph{global solution} is obtained by combining the incumbent solutions of all routing subproblems,
                \begin{equation}
                    S=\bigcup_{k=1}^{K} S_k.
                \end{equation}
            \end{definition}
            
            In the conventional CFRS framework, the decomposition $\mathcal{P}$ remains fixed throughout route construction. Therefore, each incumbent solution is computed using only the customers and vehicles assigned to its corresponding routing subproblem.
        
        \subsubsection{Evolving Formulation}
            We extend the static formulation by allowing the decomposition to evolve over a sequence of optimization steps. In this paper, an evolution step refers to a redistribution of customers or vehicles across routing subproblems. The redistribution changes the subproblem definitions as shown in Figure~\ref{fig:cfrs_dynamic}, but it does not directly modify incumbent routes whose assigned customers and vehicles remain unchanged.
            
            To make this distinction explicit, let $\mathcal{C}(S_k)=\{c\in C_k : c \text{ is served in } S_k\} $ denote the customers served by incumbent solution $S_k$, and let $ \mathcal{M}(S_k)=\{m\in M_k : R_m\in S_k\} $ denote the vehicles used in $S_k$. An incumbent subproblem solution $S_k^{(t-1)}$ can be preserved at step $t$ whenever its served customers and used vehicles remain assigned to the same subproblem,
            \begin{equation}
                \mathcal{C}(S_k^{(t-1)}) \subseteq C_k^{(t)}
                \qquad\text{and}\qquad
                \mathcal{M}(S_k^{(t-1)}) \subseteq M_k^{(t)}.
                \label{eqn:evolve}
            \end{equation}
            
            Equivalently, an incumbent route is affected only if at least one of its served customers or its associated vehicle is redistributed to another subproblem.
            
            Let
            \begin{equation}
                \mathcal{P}^{(t)}=\left\{(C_1^{(t)},M_1^{(t)}),\dots,(C_K^{(t)},M_K^{(t)})\right\}
            \end{equation}
            denote the decomposition after optimization step $t$. Each decomposition induces routing subproblems
            \begin{equation}
                X_k^{(t)}=(C_k^{(t)},M_k^{(t)}),
                \label{eqn:subproblem_dyn}
            \end{equation}
            and corresponding incumbent solutions
            \begin{equation}
                S_k^{(t)}=\sigma\!\left(X_k^{(t)},S_k^{(t-1)}\right).
                \label{eqn:solution_dyn}
            \end{equation}
            
            By a slight abuse of notation, we allow the routing procedure $\sigma$ to accept the previous incumbent solution as an additional argument. This indicates that previously computed routing information may be used when constructing the updated incumbent solution.
            
            The collection of incumbent solutions after optimization step $t$ is
            \begin{equation}
                \mathcal{S}^{(t)}=\left\{S_1^{(t)},\dots,S_K^{(t)}\right\},
            \end{equation}
            and the corresponding global solution is
            \begin{equation}
                S^{(t)}=\bigcup_{k=1}^{K}S_k^{(t)}.
            \end{equation}
            
            The optimization therefore evolves through a sequence of decompositions and incumbent solutions,
            \begin{equation}
                (\mathcal{P}^{(0)},\mathcal{S}^{(0)})
                \rightarrow
                (\mathcal{P}^{(1)},\mathcal{S}^{(1)})
                \rightarrow \cdots\rightarrow
                (\mathcal{P}^{(T)},\mathcal{S}^{(T)}).
            \end{equation}
            
            At each optimization step, subproblems whose assigned customers or vehicles change may require re-optimization, while incumbent routes whose customers and vehicles remain assigned to the same subproblem can be preserved.

    The formulation presented here establishes the mathematical framework for the proposed routing methodology. The following section describes the collaboration operations and optimization procedure that evolve the decomposition and incumbent solutions during the routing process.

\section{Collaborative Route Construction}
    Building upon the evolving CFRS formulation introduced in Section~\ref{sec:problem_formulation}, CoRC introduces an intermediate coordination stage into route building, enabling routing subproblems to exchange customers and vehicles while preserving independent optimization.

    \begin{figure*}[h]
        \centering
        \includegraphics[width=\textwidth]{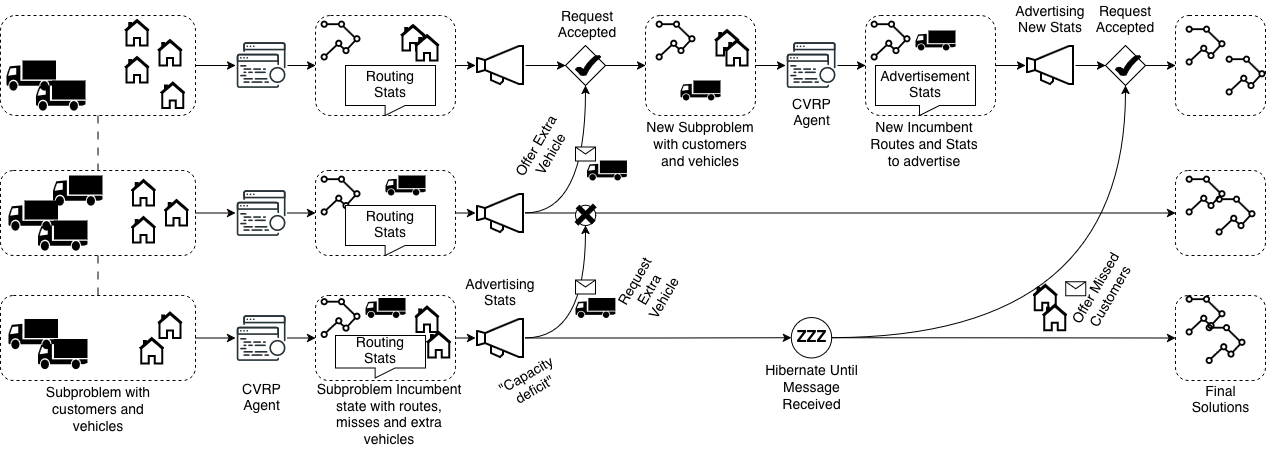}
        \caption{
        Overview of the collaborative route-construction framework. Independent routing agents solve decomposed subproblems, compute and advertise local solution statistics (such as "Capacity deficit"), exchange resources when shortages and surpluses are detected, and iteratively rebuild routes until a globally feasible solution is obtained.
        }
        \label{fig:framework}
    \end{figure*}
    
    Figure~\ref{fig:framework} illustrates the overall organization of our method. Starting from an initial decomposition, routing agents repeatedly (i) update incumbent routing solutions, (ii) derive and exchange routing statistics, and (iii) perform collaboration operations before the affected routing subproblems are re-optimized. The remainder of this section describes these three components in detail.

    \subsection{Routing State Management by Partition Agents} \label{sec:state_mgmt}
        As defined in Section~\ref{sec:problem_formulation}, each routing subproblem $X_k^{(t)}$ maintains an incumbent solution $S_k^{(t)}$, obtained through the routing procedure $\sigma(\cdot)$ in \eqref{eqn:solution_dyn}. The formulation permits $\sigma(\cdot)$ to exploit the incumbent solution from the previous optimization step but otherwise remains agnostic to the underlying CVRP solver. Thus, any routing algorithm compatible with the formulation of Section~\ref{sec:problem_formulation} may be employed within the framework.
        
        \subsubsection{Routing Statistics} \label{sec:stats}
            Following each invocation of the routing procedure, every routing agent derives a compact summary of its current routing outcome,
            \begin{equation}
                \Phi_k^{(t)}
                =
                f\!\left(
                    X_k^{(t)},
                    S_k^{(t)}
                \right),
            \end{equation}
            where $f(\cdot)$ denotes the statistics extraction procedure. The resulting statistics summarize the routing resources currently required by the routing subproblem together with the resources that remain available for collaboration.
            
            Let $U_k^{(t)}\subseteq C_k^{(t)}$ denote the set of customers that remain unserved after route construction.
            
            The smallest missed customer demand is
            \begin{equation}
                D_k^{\min} = \min_{c\in U_k^{(t)}} d_c,
            \end{equation}
            representing the smallest customer demand that remains unserved. This quantity is used to determine whether another routing subproblem possesses sufficient remaining capacity to accommodate at least one missed customer.
            
            The remaining vehicle capacity is
            \begin{equation}
                Q_k^{\mathrm{free}}
                =
                \sum_{m\in M_k^{(t)}}Q_m
                -
                \sum_{c\in C_k^{(t)}\setminus U_k^{(t)}}d_c,
            \end{equation}
            which measures the unused carrying capacity available after the incumbent routing solution has been constructed.
            
            The number of unserved customers is
            \begin{equation}
                N_k^{\mathrm{miss}}
                =
                |U_k^{(t)}|,
            \end{equation}
            while the number of vehicles that remain unused is
            \begin{equation}
                N_k^{\mathrm{free}}
                =
                |E_k^{(t)}|.
            \end{equation}
            
            Finally, each routing agent records the aggregate customer demand and available vehicle capacity within the routing subproblem,
            \begin{equation}
                D_k^{\mathrm{tot}}
                =
                \sum_{c\in C_k^{(t)}}d_c,
                \qquad
                Q_k^{\mathrm{tot}}
                =
                \sum_{m\in M_k^{(t)}}Q_m.
            \end{equation}
            
            Collectively, these quantities define the routing statistics vector
            \begin{equation}
            \Phi_k^{(t)}
            =
            \left(
            D_k^{\min},
            Q_k^{\mathrm{free}},
            N_k^{\mathrm{miss}},
            N_k^{\mathrm{free}},
            D_k^{\mathrm{tot}},
            Q_k^{\mathrm{tot}}
            \right).
            \end{equation}
            
            The routing statistics provide a compact summary of the current optimization state and are exchanged between routing subproblems during collaboration, avoiding the need to communicate complete routing solutions.
        
    \subsection{Collaboration Operations} \label{sec:collabs}
        The routing statistics determine whether the current decomposition should be revised during route building. Specifically, routing subproblems evaluate the advertised statistics of other subproblems to identify opportunities for collaboration before selecting one of the decomposition update operations introduced below.
        
        \begin{remark}[Communication Topology]
            Our study uses a fully connected communication topology in which every routing subproblem exchanges advertisements with all others. More restrictive topologies can be incorporated to reduce communication overhead; however, the fully connected topology yielded the best empirical performance in our experiments and is therefore adopted in this work.
        \end{remark}
        
        \subsubsection{Collaboration Selection}
            Each routing subproblem evaluates the advertised routing statistics and assigns a collaboration priority to every candidate subproblem. The framework is independent of the particular priority policy; in this work, we employ the following heuristic to rank candidate collaborations before selecting one of the collaboration operations introduced below.
            
            \begin{equation}
                \begin{aligned}
                s(i,j)=
                \begin{cases}
                2, &
                \begin{aligned}[t]
                &N_i^{\mathrm{miss}}>0 \\
                &{}\land\ D_i^{\min} < Q_j^{\mathrm{free}},
                \end{aligned}
                \\[0.6em]
                2, &
                \begin{aligned}[t]
                &N_j^{\mathrm{miss}}>0 \\
                &{}\land\ D_j^{\min} < Q_i^{\mathrm{free}} \\
                &{}\land\ N_i^{\mathrm{free}}>0,
                \end{aligned}
                \\[0.6em]
                1, &
                \begin{aligned}[t]
                &N_i^{\mathrm{miss}}+N_j^{\mathrm{miss}}>0 \\
                &{}\land\ D_i^{\mathrm{tot}}+D_j^{\mathrm{tot}}
                \le
                Q_i^{\mathrm{tot}}+Q_j^{\mathrm{tot}},
                \end{aligned}
                \\[0.6em]
                0, & \text{otherwise}.
                \end{cases}
                \end{aligned}
            \end{equation}

            \begin{remark}[Collaboration Ranking Metric]
                The collaboration policy used in this work represents one possible realization of the CoRC framework. More sophisticated decision policies may incorporate additional routing statistics, learned decision models, or optimization-based selection mechanisms without changing the collaboration operations or the evolving formulation.
            \end{remark}
        
        \subsubsection{Customer Transfer}
            A customer transfer is performed when routing subproblem $i$ contains unserved customers while routing subproblem $j$ possesses sufficient remaining capacity to accommodate at least the smallest unserved customer demand in $i$. Formally, if
            \begin{equation}
                N_i^{\mathrm{miss}}>0,
                \qquad
                D_i^{\min}<Q_j^{\mathrm{free}},
            \end{equation}
            CoRC transfers a subset of the unserved customers from $i$ to $j$. Let $\Delta C\subseteq U_i^{(t)}$ denote the transferred customer set. The updated customer assignments become
            \begin{equation}
                C_i^{(t+1)}
                =
                C_i^{(t)}\setminus\Delta C,
                \qquad
                C_j^{(t+1)}
                =
                C_j^{(t)}\cup\Delta C.
            \end{equation}
            
            The affected routing subproblems are subsequently re-optimized according to the evolving decomposition formulation.
            
        \subsubsection{Vehicle Transfer}
            A vehicle transfer is performed when routing subproblem $j$ contains unserved customers while routing subproblem $i$ possesses sufficient remaining capacity together with at least one unused vehicle. Formally, if
            \begin{equation}
                N_j^{\mathrm{miss}}>0,
                \qquad
                D_j^{\min}<Q_i^{\mathrm{free}},
                \qquad
                N_i^{\mathrm{free}}>0,
            \end{equation}
            CoRC transfers one or more unused vehicles from $i$ to $j$. Let $\Delta M\subseteq E_i^{(t)}$ denote the transferred vehicle set. The updated vehicle assignments are
            \begin{equation}
                M_i^{(t+1)}
                =
                M_i^{(t)}\setminus\Delta M,
                \qquad
                M_j^{(t+1)}
                =
                M_j^{(t)}\cup\Delta M.
            \end{equation}
            
            The affected routing subproblems are subsequently re-optimized using the routing procedure $\sigma(\cdot)$ as in \eqref{eqn:solution_dyn}.
            
        \subsubsection{Subproblem Merge}
            When neither customer nor vehicle transfers satisfy the collaboration conditions, the two routing subproblems are merged into a single routing problem. The merged routing subproblem is defined as
            \begin{equation}
                C_{ij}^{(t+1)}
                =
                C_i^{(t)}\cup C_j^{(t)},
                \qquad
                M_{ij}^{(t+1)}
                =
                M_i^{(t)}\cup M_j^{(t)}.
            \end{equation}
            
            The merged routing subproblem is subsequently re-optimized, while all unaffected routing subproblems retain their incumbent solutions.
        
        Accepted proposals update the affected routing subproblems, after which the corresponding incumbent solutions and routing statistics are recomputed according to the evolving decomposition formulation.
        
    \subsection{Asynchronous Agent Decision Cycle} \label{sec:cycle}
        CoRC is implemented as an asynchronous optimization cycle executed independently by each routing agent as described in Algorithm~\ref{alg:corc-agent}. At optimization step $t$, agent $k$ maintains routing subproblem $X_k^{(t)}=(C_k^{(t)},M_k^{(t)})$ and incumbent solution $S_k^{(t)}$. Whenever the customer or vehicle assignment of the subproblem changes, the agent invokes the routing operator $\sigma$ to update its incumbent solution and then recomputes the statistics vector $\Phi_k^{(t)}$. These statistics are advertised to the other agents and are also used to evaluate received advertisements.
        
        The asynchronous structure allows that agents do not wait for a global synchronization round before acting. Instead, each agent alternates between local route construction, advertisement publication, advertisement reading, and proposal validation. Sleep parameters are included to avoid continuous polling and to control the frequency of communication and proposal processing. These parameters, together with the choice of routing operator $\sigma$, are reported in the computational settings.

        An agent remains active until either the global time budget is exhausted or all of its customers and vehicles have been transferred to another routing subproblem. In the latter case, the agent reaches an absorbing state and no longer participates in the optimization process. Hence, successive merge operations produce progressively larger routing subproblems that accumulate customers, vehicles, and incumbent routing information over time. Rather than performing a single global re-optimization after all independent routing has been completed, as in decomposition-based approaches such as \citep{kerscher2024spatial}, CoRC incrementally increases the optimization scope through localized merge operations. This allows larger routing subproblems to emerge only where collaboration indicates additional joint optimization may be beneficial, while unaffected routing subproblems continue to evolve independently.
        
        \begin{algorithm}[h]
            \caption{Asynchronous CoRC Optimization Cycle for Agent $k$}
            \label{alg:corc-agent}
            \begin{algorithmic}[1]
            \Require Initial subproblem $X_k^{(0)}=(C_k^{(0)},M_k^{(0)})$, initial incumbent $S_k^{(-1)}$, routing operator $\sigma$, end time $T$, advertisement wait time $\tau_{\mathrm{adv}}$, proposal wait time $\tau_{\mathrm{prop}}$
            \State $t \gets 0$
            \While{current time $< T$}
                \If{state has changed, i.e., $X_k^{(t)} \neq X_k^{(t-1)}$}
                    \State Invoke routing operator, $S_k^{(t)} \gets \sigma\!\left(X_k^{(t)},S_k^{(t-1)}\right)$
                    \State Recompute statistics, $\Phi_k^{(t)} \gets f\!\left(X_k^{(t)},S_k^{(t)}\right)$
                \Else
                    \State Retain $S_k^{(t)} \gets S_k^{(t-1)}$
                    \State Retain $\Phi_k^{(t)} \gets \Phi_k^{(t-1)}$
                \EndIf
            
                \State Publish advertisement $\Phi_k^{(t)}$ to other agents
                \State Sleep for $\tau_{\mathrm{adv}}$ duration
                \State Read advertisements from other agents $\{\Phi_\ell^{(t_\ell)}:\ell\neq k\}$
            
                \State Find the highest candidate, $j^\star \gets \arg\max_{j\neq k} s(k,j)$
                \If{$s(k,j^\star)>0$}
                    \State Generate proposal $a\in\{\textsc{CustomerTransfer},\textsc{VehicleTransfer},\textsc{Merge}\}$
                    \State Sleep for $\tau_{\mathrm{prop}}$ duration
                    \If{$a$ is valid under current advertisements}
                        \State Apply $a$ to update affected subproblems
                        \State $X_k^{(t+1)} \gets (C_k^{(t+1)},M_k^{(t+1)})$
                    \EndIf
                \EndIf
            
                \If{no customers and no vehicles in the partition}
                    \State \textbf{break}
                \EndIf
                \State $t \gets t+1$
            \EndWhile
            \State \Return $S_k^{(t)}$
            \end{algorithmic}
        \end{algorithm}

        The proposed framework exposes a small number of configurable parameters that determine the behavior of the optimization cycle. These include the routing operator $\sigma$, the advertisement and proposal waiting times $(\tau_{\mathrm{adv}},\tau_{\mathrm{prop}})$, and the overall optimization time budget. The waiting times regulate the frequency of inter-agent communication and proposal evaluation, balancing communication overhead against responsiveness in the asynchronous optimization process. Table~\ref{tab:corc_parameters} summarizes the parameter values used throughout the computational experiments. Unless stated otherwise, these values were selected empirically based on preliminary tuning and were held fixed across all benchmark instances.

        \begin{table}[t]
            \centering
            \caption{CoRC parameters used in the computational experiments.}
            \label{tab:corc_parameters}
            \begin{tabular}{lll}
            \hline
            \textbf{Parameter} & \textbf{Description} & \textbf{Value} \\
            \hline
            $\sigma$ & Routing operator & OR-Tools GLS \citep{ortools_routing} \\
            $\tau_{\mathrm{adv}}$ & Advertisement wait time & 5,000 ms \\
            $\tau_{\mathrm{prop}}$ & Proposal wait time & 200 ms \\
            $T$ & Optimization time budget & Instance/Experiment dependent \\
            \hline
            \end{tabular}
        \end{table}

\section{Experiments and Evaluation}
    This section describes the computational study conducted to evaluate the CoRC framework. We first introduce the experimental methodology, including the evaluation metrics and comparison methods, and then present a sequence of experiments designed to assess the effectiveness, scalability, and robustness of the collaboration mechanism.

    \subsection{Experiment Setup}
        \subsubsection{Evaluation Metrics}
            We evaluate our study using three performance measures. The primary metric is the \emph{time to full solution}, defined as the elapsed time required to construct a feasible solution in which all customer demand is served. For instances where multiple methods reach a full solution, we additionally report the corresponding routing distance. When a method does not produce a full solution within the prescribed time limit, its progress is quantified using the percentage of customer demand that remains unserved,
            
            \begin{equation}
                \mathrm{MissedDemand}(\%)=100\cdot\frac{\sum_{c\in U} d_c}{\sum_{c\in C} d_c},
            \end{equation}
            where $U$ denotes the set of unserved customers. These metrics jointly evaluate solution quality, convergence speed, and progress toward complete feasibility.

            For methods that could not complete due to computational resource limitations (e.g., memory exhaustion) are reported as N/A, whereas methods that completed execution but failed to construct a full solution within the prescribed time budget are reported as INF in the results tables.
    
        \subsubsection{Clustering Methods} \label{sec:clustering}
            To evaluate the proposed collaboration framework under diverse decomposition strategies, we consider four representative clustering methods. \textbf{Sweep} \citep{gillett1974heuristic} and \textbf{Grid} generate spatially uniformly split rectangular boundaries. \textbf{CC-CVRP} \citep{alesiani2022constrained} builds routing subproblems using demand and capacity information during decomposition. Finally, \textbf{Random} assigns customers to routing subproblems without considering either spatial or operational structure, providing a challenging reference decomposition.

            Unless the clustering algorithm outlines the vehicle distribution mechanism explicitly, we distribute the vehicles in a balanced manner so that clusters with higher total demand receive higher total vehicle capacity, proportionally.
            
            These methods span a broad spectrum of decomposition characteristics, allowing the collaboration mechanism to be evaluated independently of any particular clustering strategy.
    
        \subsubsection{Compared Methods}
            The evaluated methods are grouped into four categories.
    
            The first category consists of independent optimization frameworks, denoted by the suffix \textbf{-Ind}. For each clustering method, routing subproblems are solved independently using the common routing operator $\sigma$, after which the resulting routes are combined to form the final routing solution.
            
            The second category consists of global re-optimization baseline, denoted by the suffix \textbf{-Reopt}. This method first solves the decomposed routing subproblems independently, as in the \textbf{-Ind} baselines. The only difference is that the independent solvers are terminated when the routing solver determines no more customers can be served, or half of the total allotted time is reached. The combined solution is then used as the initial solution for a global optimization step, which is run for the remaining time budget of 6 hours.
            
            The third category applies the CoRC framework to the same decompositions. Since the independent, global re-optimization, and CoRC variants share identical decompositions and routing operators, these comparisons isolate the effect of when coordination is introduced during the routing process. We label these runs with the suffix \textbf{-CoRC}. All methods under independent, re-optimization, and the CoRC use Guided Local Search \citep{voudouris1999guided} in Or-Tools implementation \citep{ortools_routing}.
            
            The fourth category consists of complete routing frameworks that do not explicitly separate decomposition and routing. We consider \textbf{ScaleNet} \citep{liu2026scale}, a learning-based routing framework, and \textbf{Hybrid Genetic Search (HGS)} \citep{vidal2022hybrid}, one of the strongest general-purpose metaheuristics for the CVRP.

    \subsection{Performance on Benchmark and Large-Scale Instances}
        Our first computational study evaluates the overall routing performance of the CoRC framework across both established benchmark instances and large-scale synthetic instances. The objective is to compare CoRC against independent routing, post-routing global re-optimization, and complete routing frameworks under a common experimental setting while assessing its behavior across a broad range of problem sizes.

        The study consists of two datasets. The first is the AGS benchmark suite \citep{arnold2019efficiently} comprising the Ghent1 (10K), Brussels1 (15K), Flanders1 (20K), and Flanders2 (30K) instances. The second consists of synthetic instances containing 100,000 and 200,000 customers. Customer locations are sampled uniformly over the service region. Vehicle capacities are drawn uniformly from the interval $[2d_{\max},5d_{\max}]$, where $d_{\max}$ denotes the maximum customer demand. The fleet size is fixed to 400 vehicles for both synthetic instances, and the generated capacities are normalized to obtain an overall capacity utilization of $0.95$.

        The computational budget is set to one hour for the AGS benchmark instances and six hours for the large-scale synthetic instances, due to the significant difference in problem sizes. Results for the AGS and synthetic datasets are reported in Tables~\ref{tab:ags_results} and~\ref{tab:large_results}, respectively.

        \begin{table*}[t]
            \centering
            \caption{Comparison of decomposition-based and complete methods on the AGS benchmark instances. Runs that did not obtain a full solution within the 6-hour limit are denoted ``INF'' and experiments failing due to excessive memory are ''N/A''.}
            \label{tab:ags_results}
            
            \begin{subtable}[t]{0.49\textwidth}
                \centering
                \caption{Ghent1 (10K)}
                \resizebox{\linewidth}{!}{
                \begin{tabular}{lrrr}
                \toprule
                \textbf{Method} &
                \textbf{Time (s)} &
                \textbf{Distance} &
                \textbf{Missed (\%)}\\
                \midrule
                
                Sweep-Ind        & INF   &   714,447 & 1.95 \\
                Sweep-Reopt      & INF   &   520,747 & 2.12 \\
                Sweep-CoRC       & 0.04  & 2,686,808 & 0.00 \\
                \addlinespace
                
                Grid-Ind         & INF   &   428,595 & 47.69 \\
                Grid-Reopt       & INF   &   330,197 & 49.16 \\
                Grid-CoRC        & 0.04  & 1,644,217 & 0.00 \\
                \addlinespace
                
                Random-Ind       & INF   & 1,799,392 & 1.92 \\
                Random-Reopt     & INF   &   957,593 & 1.86 \\
                Random-CoRC      & 0.04  & 5,556,353 & 0.00 \\
                \addlinespace
                
                CC-CVRP-Ind      & INF   &   577,955 & 3.11 \\
                CC-CVRP-Reopt    & INF   &   495,373 & 1.68 \\
                CC-CVRP-CoRC     & 10.42 & 1,515,806 & 0.00 \\
                
                \midrule
                
                ScaleNet         & 18,176.20 &   469,531 & 0.00 \\
                HGS              &    833.77 &   515,551 & 0.00 \\
                
                \bottomrule
                \end{tabular}}
            \end{subtable}
            \hfill
            \begin{subtable}[t]{0.49\textwidth}
                \centering
                \caption{Instance: Brussels1 (15K)}
                \resizebox{\linewidth}{!}{
                \begin{tabular}{lrrr}
                \toprule
                \textbf{Method} &
                \textbf{Time (s)} &
                \textbf{Distance} &
                \textbf{Missed (\%)}\\
                \midrule
                
                Sweep-Ind        & INF   & 1,393,274 & 1.77 \\
                Sweep-Reopt      & INF    & 646,201   & 2.03 \\
                Sweep-CoRC       & 0.04  & 4,272,195 & 0.00 \\
                \addlinespace
                
                Grid-Ind         & INF   & 800,341   & 42.46 \\
                Grid-Reopt       & INF    & 418,720   & 43.70 \\
                Grid-CoRC        & 0.04  & 2,354,391 & 0.00 \\
                \addlinespace
                
                Random-Ind       & INF   & 3,884,946 & 1.77 \\
                Random-Reopt     & INF    & 1,279,333 & 2.03 \\
                Random-CoRC      & 0.04  & 8,958,221 & 0.00 \\
                \addlinespace
                
                CC-CVRP-Ind      & INF   & 939,889   & 2.13 \\
                CC-CVRP-Reopt    & INF    & 541,737   & 3.40 \\
                CC-CVRP-CoRC     & 15.42 & 2,160,617 & 0.00 \\
                
                \midrule
                ScaleNet         & 22,656.22 & 1,253,335 & 0.00 \\
                HGS & 469.61    & 567,312   & 0.00 \\
                
                \bottomrule
                \end{tabular}}
            \end{subtable}
            
            \vspace{0.8em}
            
            \begin{subtable}[t]{0.49\textwidth}
                \centering
                \caption{Instance: Flanders1 (20K)}
                \resizebox{\linewidth}{!}{
                \begin{tabular}{lrrr}
                \toprule
                \textbf{Method} &
                \textbf{Time (s)} &
                \textbf{Distance} &
                \textbf{Missed (\%)}\\
                \midrule
                Sweep-Ind        & INF   & 21,984,887  & 7.49 \\
                Sweep-Reopt      & INF    & 7,710,955  & 9.61 \\
                Sweep-CoRC       & 0.05  & 53,668,458  & 0.00 \\
                \addlinespace
                
                Grid-Ind         & INF   & 11,609,403  & 35.79 \\
                Grid-Reopt       & INF    & 6,743,007  & 37.33 \\
                Grid-CoRC        & 0.05  & 26,310,823  & 0.00 \\
                
                \addlinespace
                Random-Ind       & INF   & 69,229,907  & 1.08 \\
                Random-Reopt     & INF    & 16,682,198 & 1.28 \\
                Random-CoRC      & 0.05  & 139,684,941 & 0.00 \\
                
                \addlinespace
                CC-CVRP-Ind      & INF   & 12,363,983  & 9.38 \\
                CC-CVRP-Reopt    & INF    & 6,412,808  & 11.85 \\
                CC-CVRP-CoRC     & 20.08 & 31,615,917  & 0.00 \\
                
                \midrule
                ScaleNet         & 15,712.07 & 7,240,118 & 0.00 \\
                HGS              & 786.27   & 7,858,437   & 0.00 \\
                \bottomrule
                \end{tabular}}
            \end{subtable}
            \hfill
            \begin{subtable}[t]{0.49\textwidth}
                \centering
                \caption{Instance: Flanders2 (30K)}
                \resizebox{\linewidth}{!}{
                \begin{tabular}{lrrr}
                \toprule
                \textbf{Method} &
                \textbf{Time (s)} &
                \textbf{Distance} &
                \textbf{Missed (\%)}\\
                \midrule
                
                Sweep-Ind        & INF   & 44,560,865   & 8.19 \\
                Sweep-Reopt      & INF    & 7,523,698  & 8.39 \\
                Sweep-CoRC       & 0.04  & 47,103,331   & 0.00 \\
                \addlinespace
                
                Grid-Ind         & INF   & 14,364,824   & 37.58 \\
                Grid-Reopt       & INF    & 6,026,707  & 38.04 \\
                Grid-CoRC        & INF   & 31,618,713   & 9.38 \\
                \addlinespace
                
                Random-Ind       & INF   & 151,233,810  & 1.59 \\
                Random-Reopt     & INF    & 25,411,542 & 4.03 \\
                Random-CoRC      & 0.04  & 202,791,738  & 0.00 \\
                \addlinespace
                
                CC-CVRP-Ind      & INF   & 21,213,338   & 8.37 \\
                CC-CVRP-Reopt    & INF   & 5,553,576  & 9.33 \\
                CC-CVRP-CoRC     & 20.07 & 35,773,046   & 0.00 \\
                
                \midrule
                
                ScaleNet         & 32,581.33 & 4,373,244 & 0.00 \\
                HGS              & 3,043.79 & 5,154,561   & 0.00 \\
                
                \bottomrule
                \end{tabular}}
            \end{subtable}
        \end{table*}
    
        \begin{table*}[t]
            \centering
            \caption{Comparison of decomposition-based and complete routing methods on large synthetic instances. Runs that did not obtain a full solution within the 6-hour limit are denoted ``INF'' and experiments failing due to excessive memory are ''N/A''.}
            \label{tab:large_results}
            
            \begin{subtable}[t]{0.49\textwidth}
                \centering
                \caption{Gen100K}
                \resizebox{\linewidth}{!}{
                \begin{tabular}{lrrr}
                \toprule
                \textbf{Method} &
                \textbf{Time (s)} &
                \textbf{Distance} &
                \textbf{Missed (\%)}\\
                \midrule
                
                Sweep-Ind        & INF   &    38,751,662 & 88.49 \\
                Sweep-Reopt      & INF   & 1,613,939,369 & 1.75 \\
                Sweep-CoRC       & 0.04  & 1,778,882,015 & 0.00 \\
                \addlinespace
                
                Grid-Ind         & INF   &    94,889,835 & 77.25 \\
                Grid-Reopt       & INF   &   327,742,615 & 51.59 \\
                Grid-CoRC        & 0.05  & 1,145,379,681 & 0.00 \\
                \addlinespace
                
                Random-Ind       & INF   &   101,295,538 & 89.55 \\
                Random-Reopt     & INF & 7,996,921,805 & 0.11 \\
                Random-CoRC      & 0.04  & 8,564,687,063 & 0.00 \\
                \addlinespace
                
                CC-CVRP-Ind      & INF   &    29,420,504 & 88.92 \\
                CC-CVRP-Reopt    & INF & 1,083,989,629 & 1.42 \\
                CC-CVRP-CoRC     & 20.15 & 1,339,605,849 & 0.00 \\
                
                \midrule
                ScaleNet         & N/A & N/A & 100.00 \\
                HGS              & N/A & N/A & 100.00 \\
                
                \bottomrule
                \end{tabular}}
            \end{subtable}
            \hfill
            \begin{subtable}[t]{0.49\textwidth}
                \centering
                \caption{Gen200K}
                \resizebox{\linewidth}{!}{
                \begin{tabular}{lrrr}
                \toprule
                \textbf{Method} &
                \textbf{Time (s)} &
                \textbf{Distance} &
                \textbf{Missed (\%)}\\
                \midrule
                
                Sweep-Ind        & INF   &    59,121,323 & 96.25 \\
                Sweep-Reopt      & INF & 2,116,206,210 & 56.17 \\
                Sweep-CoRC       & 0.06  & 7,519,598,597 & 0.00 \\
                \addlinespace
                
                Grid-Ind         & INF   &    70,788,586 & 93.42 \\
                Grid-Reopt       & INF   & 2,094,541,196 & 56.80 \\
                Grid-CoRC        & 0.07  & 4,224,960,278 & 0.00 \\
                \addlinespace
                
                Random-Ind       & INF   &   249,427,628 & 96.35 \\
                Random-Reopt     & INF & 11,809,494,910 & 55.44 \\
                Random-CoRC      & 0.05  &39,980,572,554 & 0.00 \\
                \addlinespace
                
                CC-CVRP-Ind      & INF   &    60,971,185 & 96.18 \\
                CC-CVRP-Reopt    & INF   & 16,754,947,193 & 42.79 \\
                CC-CVRP-CoRC     & 20.19 & 5,474,016,941 & 0.00 \\
                
                \midrule
                ScaleNet         & N/A & N/A & 100.00 \\
                HGS              & N/A & N/A & 100.00 \\
                
                \bottomrule
                \end{tabular}}
            \end{subtable}
        \end{table*}

        Tables~\ref{tab:ags_results} and~\ref{tab:large_results} demonstrate that CoRC is the only decomposition-based routing framework that consistently constructs full solutions across all AGS benchmark instances and both synthetic large-scale instances. In contrast, neither Independent routing nor post-routing global re-optimization reaches a full solution under any decomposition. Although the global re-optimization stage generally reduces the amount of unserved demand relative to Independent routing, substantial portions of customer demand remain unserved, particularly on the synthetic instances. These results indicate that introducing coordination during route construction is considerably more effective at recovering feasibility than postponing coordination until a subsequent global optimization stage.

        The routing distances further demonstrate that CoRC complements rather than replaces the underlying decomposition strategy. While collaboration consistently restores feasibility, the resulting routing distances continue to reflect the quality of the initial partitioning. In particular, Random decomposition produces substantially longer routes despite reaching full solutions, whereas Grid and CC-CVRP generally yield shorter routes. This behavior suggests that decomposition quality primarily determines route efficiency, while the proposed collaboration mechanism improves the ability to yield complete solutions across diverse decomposition strategies.

        On the AGS benchmark instances, the complete routing frameworks ScaleNet and HGS generally produce shorter routing distances than the decomposition-based methods, as expected from optimizing the original routing problem directly. However, neither framework successfully produced routing solutions for the 100K or 200K synthetic instances within the available computational resources, whereas CoRC continued to obtain full routing solutions across all evaluated decompositions. These results suggest that collaborative partition optimization provides a practical complement to decomposition-based routing, extending its applicability to substantially larger problem instances while preserving the flexibility of the underlying decomposition strategy.

    \subsection{Sensitivity and Robustness Analysis}
        The previous experiments demonstrated the overall effectiveness of CoRC, but they did not isolate the impact of the initial decomposition method. In this section, we evaluate the robustness of the framework under increasingly challenging partitioning conditions. To provide a controlled setting, all experiments use grid partitioning, which does not exploit spatial or demand information and therefore represents one of the least balanced clustering strategies, offering a challenging benchmark for collaboration.

        \begin{figure*}[t]
            \centering
            \begin{subfigure}[t]{0.49\textwidth}
                \centering
                \includegraphics[width=\linewidth]{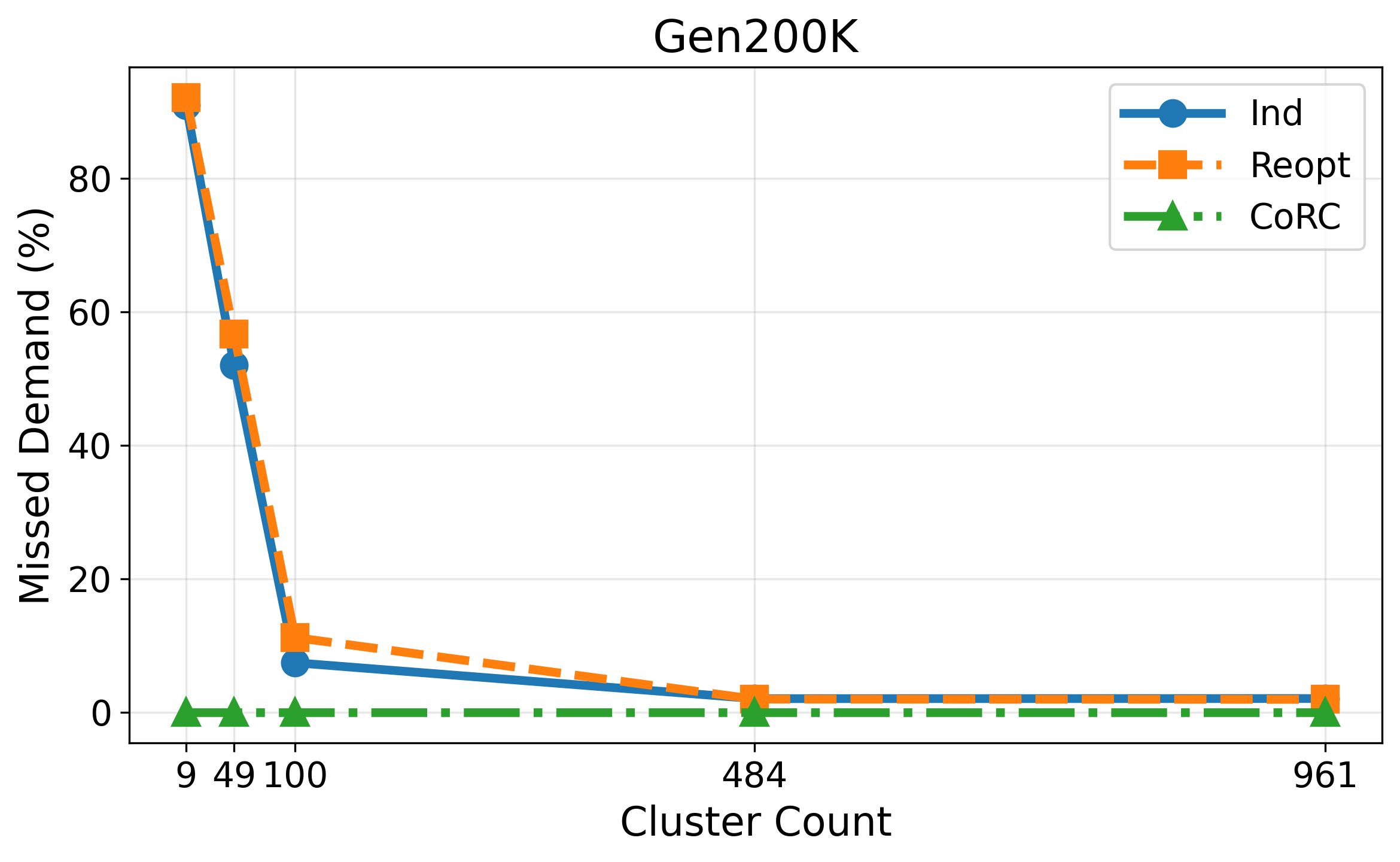}
                \caption{Influence of routing subproblem granularity on solution quality.}
                \label{fig:cluster_count}
            \end{subfigure}
            \hfill
            \begin{subfigure}[t]{0.49\textwidth}
                \centering
                \includegraphics[width=\linewidth]{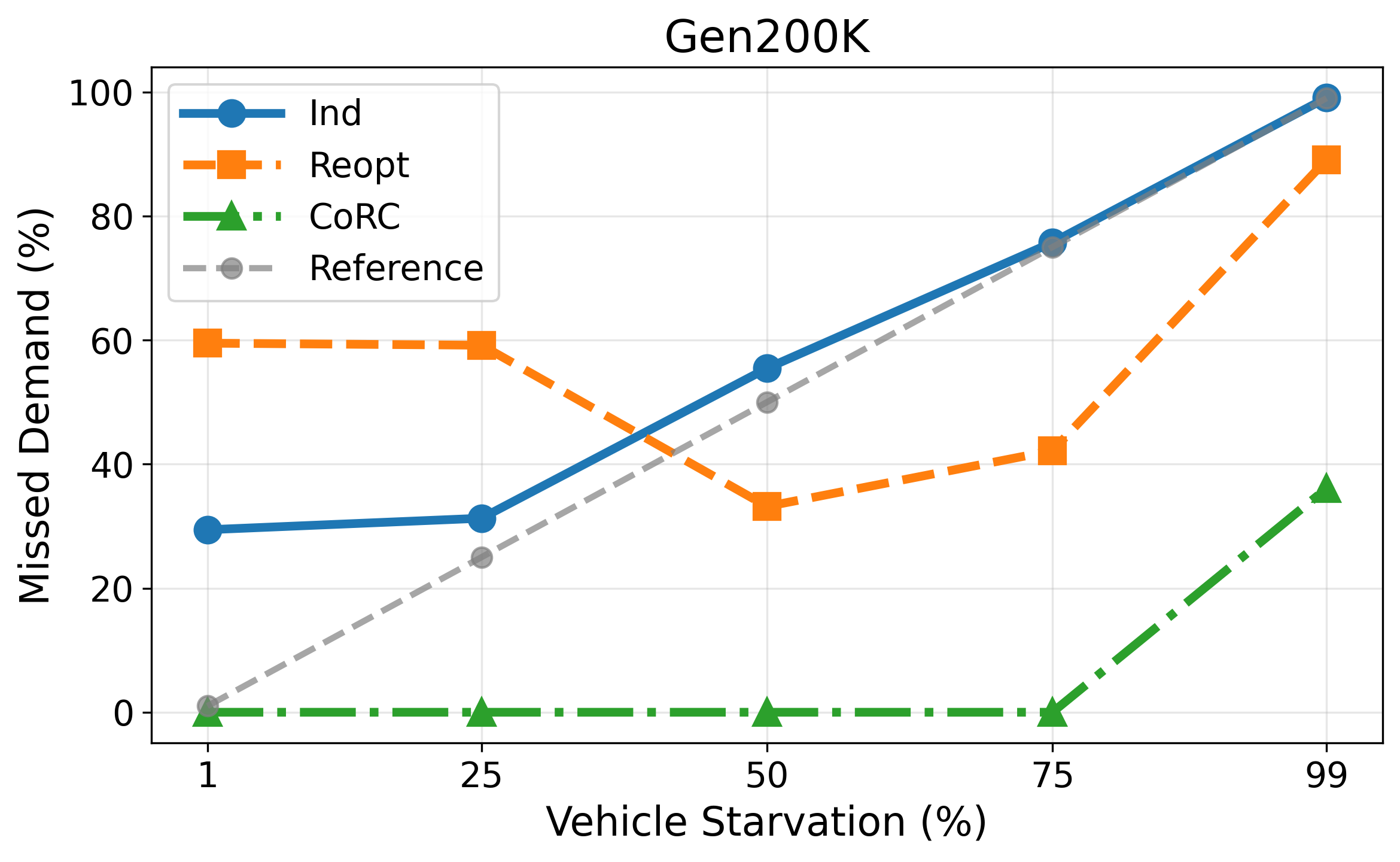}
                \caption{Impact of increasing vehicle starvation on the missed demand of frameworks. The dashed reference line denotes the starvation level ($y=x$).}
                \label{fig:vehicle_starvation}
            \end{subfigure}
            \caption{Sensitivity of the three routing strategies to decomposition granularity and vehicle imbalance.}
            \label{fig:sensitivity}
        \end{figure*}

    \subsubsection{Sensitivity to Decomposition Granularity}
        In this experiment, we investigate the sensitivity of the collaboration method to the granularity of the decomposition itself. Specifically, we study how the number of routing subproblems influences the behavior of independent routing, global re-optimization, and CoRC.
        
        The experiment is conducted on the Gen200K synthetic instance using the \textbf{Grid} decomposition method. The number of routing subproblems is varied over $\{9, 49, 100, 484, 961\}$ (all have integer square-roots and are closest to $ \{10, 50, 100, 500, 1000\} $, respectively) while all remaining experimental settings, including the routing operator $\sigma$, remain unchanged. For each decomposition, we evaluate the three routing strategies introduced previously: independent routing (\textbf{-Ind}), global re-optimization (\textbf{-Reopt}), and the CoRC.
        
        Unlike the benchmark and scalability studies, complete routing frameworks are not included in this experiment because the objective is to isolate the effect of decomposition granularity within decomposition-based routing methods.

        Figure~\ref{fig:cluster_count} evaluates the sensitivity of the routing strategies to the number of routing subproblems. As the number of clusters increases, the missed demand decreases for all methods, indicating that finer decompositions produce routing subproblems that are easier to solve independently. At very coarse decompositions (9 and 49 clusters), both Independent and Reopt fail to yield feasible solutions, leaving more than half of the demand unserved. In contrast, CoRC consistently recovers a feasible solution with zero missed demand, demonstrating that collaboration can effectively compensate for challenging initial decompositions.

        The advantage of collaboration gradually diminishes as the decomposition becomes finer. At 484 and 961 clusters, Independent and Reopt also achieve low missed demand, leaving little opportunity for collaborative repair. This behavior indicates that CoRC primarily improves robustness rather than benefiting from any particular decomposition granularity. While the independent approaches rely on carefully selecting an appropriate number of routing subproblems, CoRC maintains high solution quality across a broad range of cluster counts, substantially reducing the sensitivity of the overall framework to decomposition design.

    \subsubsection{Robustness to Resource-Imbalanced Decompositions}
        Here, we investigate how the collaboration model behaves under increasingly imbalanced customer and vehicle assignments. The objective is to evaluate the ability of the routing strategies to adapt when the initial decomposition intentionally creates routing subproblems with highly uneven resource availability.
        
        The experiment is conducted on the Gen200K synthetic instance (200K customers) using the Grid decomposition with 100 routing subproblems. Starting from a balanced decomposition, an increasing fraction of routing subproblems is selected to receive no vehicles, while their vehicles are redistributed among the remaining subproblems, preserving the overall fleet capacity. Five imbalance scenarios are considered, where $ 1\% $, $25\%$, $50\%$, $75\%$, and $99\%$ of the routing subproblems are initialized without assigned vehicles. The three routing strategies (\textbf{Ind}, \textbf{Reopt}, and \textbf{CoRC}) are then evaluated under identical routing and optimization settings.
        
        This experiment isolates the effect of resource imbalance while keeping the decomposition method, routing operator, and overall fleet resources fixed. Consequently, differences in performance can be attributed to the ability of each routing strategy to adapt to increasingly uneven customer and vehicle assignments.

        Figure~\ref{fig:vehicle_starvation} evaluates the robustness of the routing strategies under progressively increasing levels of vehicle starvation. Independent routing exhibits an almost linear increase in missed demand, closely tracking the imposed starvation level, indicating that isolated partitions have little ability to compensate for missing fleet resources. Re-optimization provides partial improvements at moderate starvation levels but displays a non-monotonic trend, with its best performance occurring under intermediate starvation. This behavior suggests that re-optimization is challenged at both extremes: when starvation is low, the independently produced cluster solutions are already of high quality, making the time spent on global re-optimization less effective than further local optimization; conversely, under severe starvation, the partition-level infeasibility becomes too substantial to repair within the remaining time budget. 
        
        In contrast, CoRC maintains complete feasibility for starvation levels up to 75\% by proactively redistributing vehicles and customers between partitions before routing. Even under the extreme 99\% starvation scenario, CoRC serves approximately two-thirds of the total demand, substantially outperforming both independent routing and re-optimization. These results demonstrate that collaboration not only improves routing performance but also makes partition-based routing considerably more robust to severe resource imbalance.
    
\section{Conclusion}
    We presented Collaborative Routing Constructors (CoRC), a collaborative routing framework that introduces an intermediate coordination stage into cluster-first route-second solution methods for the CVRP. By allowing routing subproblems to exchange customers and vehicles during route construction, CoRC complements existing decomposition strategies without modifying the underlying routing solver. Computational experiments on AGS benchmark instances and synthetic instances containing up to 200,000 customers showed that CoRC consistently constructs full routing solutions where independent routing and post-routing global re-optimization do not, while remaining effective at scales where the evaluated complete routing frameworks did not produce solutions under the same computational budget. These findings demonstrate that collaborative partition optimization provides a robust approach to feasible large-scale routing and motivate further research into optimization frameworks that dynamically adapt decompositions during route construction.

\section{Code and Data Disclosure}\label{sec:Code and Data Disclosure}The code and data to support the numerical experiments in this paper can be requested from the corresponding author.

%\THEEndNotes
% \begingroup \parindent 0pt \parskip 0.0ex \def\enotesize{\normalsize} \theendnotes \endgroup

% Appendix here
% Options are (1) APPENDIX (with or without general title) or
%             (2) APPENDICES (if it has more than one unrelated sections)
% Outcomment the appropriate case if necessary
%
% \begin{APPENDIX}{<Title of the Appendix>}
% \end{APPENDIX}
%
%   or
%
% \begin{APPENDICES}
% \section{<Title of Section A>}
% \section{<Title of Section B>}
% etc
% \end{APPENDICES}

% Acknowledgments here
% \ACKNOWLEDGMENT{We would like to express our sincere gratitude to [acknowledge individuals, organizations, or institutions] for their invaluable contributions to this research. We are also grateful to [mention any additional acknowledgements, such as technical assistance, data providers, or colleagues] for their support and assistance throughout the course of this work.}

\bibliography{references}{}
\bibliographystyle{informs2014}

%%%%%%%%%%%%%%%%%
\end{document}